\documentclass[journal]{IEEEtran}

\usepackage{amsmath,amssymb}
\usepackage{graphicx}
\usepackage{booktabs}
\usepackage{cite}
\usepackage{xcolor}
\usepackage{hyperref}
\definecolor{lightgreenbg}{RGB}{230, 255, 230}
\usepackage{multirow}

\newcommand{\textTargetSet}{\mathbf{T}}
\newcommand{\rgbImage}[1]{\mathbf{I}^{\mathrm{rgb}}_{#1}}
\newcommand{\observ}[2]{\mathbf{O}^{#1}_{#2}}
\newcommand{\regionZones}[1]{\mathbf{Z}^{#1}}
\newcommand{\regionCurrent}[2]{z_{\mathrm{prox},#2}^{#1}}
\newcommand{\regionDist}[1]{\mathbf{Z}_{\mathrm{dist}}^{#1}}
\newcommand{\actionSpace}[1]{\mathbf{A}_{#1}}
\newcommand{\viewpoints}[1]{\mathbf{V}^{#1}}
\newcommand{\targetProb}[3]{p_{#2, #3}^{\mathrm{goal},#1}}
\newcommand{\vpAssigns}[2]{\mathbf{\Pi}_{#2}^{#1}}
\newcommand{\nodeVisitTimes}[2]{\tau_{#2}^{#1}}
\newcommand{\VPEdges}[1]{\mathbf{E}^{#1}}
\newcommand{\VPEdgesGrounded}[1]{\mathbf{E}^{#1}_{+}}
\newcommand{\VPEdgesUngrounded}[1]{\mathbf{E}^{#1}_{-}}
\newcommand{\travelDist}[2]{d_{#2}^{#1}}
\newcommand{\travelDistVar}[2]{(\sigma_{#2}^{#1})^2}
\newcommand{\edgeWrap}[2]{\{#1,#2\}}
\newcommand{\orEdgeWrap}[2]{(#1,#2)}
\newcommand{\graph}[1]{\mathbf{G}^{#1}}
\newcommand{\orCurrentVp}[1]{n_{#1}}
\newcommand{\orNodes}[1]{\mathbf{N}^{#1}}
\newcommand{\arcSelectVar}[3]{x_{#2,#3}^{#1}}
\newcommand{\routingArcs}[1]{\mathcal{A}^{#1}}
\newcommand{\evalViewpoints}[1]{\mathbf{V}_{#1}}

\newcommand{\pathLenPenalWeight}{w^{\mathrm{dist}}}
\newcommand{\arcUnreliaWeight}{w^{\mathrm{arc}}}
\newcommand{\nodeRevisitWeight}{w^{\mathrm{visit}}}

\newcommand{\pathLenPenalWeightUB}{U^\mathrm{dist}}
\newcommand{\pathLenPenalWeightLB}{L^\mathrm{dist}}
\newcommand{\arcUnreliaWeightUB}{U^\mathrm{arc}}
\newcommand{\arcUnreliaWeightLB}{L^\mathrm{arc}}
\newcommand{\nodeRevisitWeightUB}{U^\mathrm{visit}}
\newcommand{\nodeRevisitWeightLB}{L^\mathrm{visit}}

\newcommand{\agentSet}{\mathbf{Q}}
\newcommand{\targetSet}{\mathbf{K}}
\newcommand{\zoneOfVP}[2]{z_{#2}^{#1}}
\newcommand{\targetFound}[2]{f_{#2}^{#1}}
\newcommand{\multiAgentNodeSelectVar}[3]{y_{#2,#3}^{#1}}
\newcommand{\multiAgentSubtourVar}[3]{u_{#2,#3}^{#1}}
\newcommand{\MLLMcondEdgeExistProb}[2]{q_{#2}^{\mathrm{cond},*#1}}
\newcommand{\MLLMEdgeDist}[2]{d_{#2}^{*#1}}
\newcommand{\MLLMdistVarVV}[2]{(\sigma_{#2}^{*#1})^2}
\newcommand{\condEdgeExistProb}[2]{p_{\mathrm{exist},#2}^{#1,\mathrm{cond}}}
\newcommand{\priorCondEdgeExistProb}[2]{\bar q_{#2}^{#1}}
\newcommand{\edgeCompScore}[2]{c^{#1}_{#2}}
\newcommand{\edgeExistLikelihood}[2]{\lambda_{#2}^{#1,+}}
\newcommand{\edgeNonExistLikelihood}[2]{\lambda_{#2}^{#1,-}}
\newcommand{\smallNumber}{\varepsilon}
\newcommand{\vpAssignIndicator}[2]{v_{#2}^{#1}}
\newcommand{\empMeanEdgeDist}[1]{\mu_{\mathrm{vv}}^{#1}}
\newcommand{\empVarEdgeDist}[1]{(\sigma_{\mathrm{vv}}^{#1})^2}
\newcommand{\edgeExistLikelihoodFactor}{\varrho}
\newcommand{\edgeExistLatent}[1]{\widehat{e}_{#1}}
\newcommand{\edgeDistLatent}[1]{\delta_{#1}}
\newcommand{\edgeDistPriorMean}[2]{\bar d_{#2}^{#1}}
\newcommand{\edgeDistPriorVar}[2]{(\bar \sigma_{#2}^{#1})^2}
\newcommand{\cueEdgeDist}[2]{\tilde d_{#2}^{#1}}
\newcommand{\cueEdgeDistVar}[2]{(\tilde \sigma_{#2}^{#1})^2}
\newcommand{\VVedgeCueFactor}{\kappa_{\mathrm{vv}}}

\newcommand{\eligibleVPs}[1]{\mathbf{V}^{#1}_{\mathrm{elig}}}
\newcommand{\rawVPGoalScore}[3]{\xi^{#1}_{#2,#3}}
\newcommand{\actTargets}[1]{\targetSet_{\mathrm{act}}^{#1}}
\newcommand{\obsVVEdges}[1]{\overline{\VPEdges{#1}}}
\newcommand{\deadEndNodes}[1]{\mathcal{B}^{#1}}

\newcommand{\currentVPSet}[1]{\mathbf{C}^{#1}}
\newcommand{\rewardVPs}[2]{\mathbf{R}_{#2}^{#1}}
\newcommand{\candidateNodes}[2]{\mathbf{N}_{#2}^{#1}}
\newcommand{\targetAssignVar}[4]{a_{#2,#3,#4}^{#1}}
\newcommand{\firstHopEdges}[2]{\mathbf{F}_{#2}^{#1}}

\newcommand{\normTerm}[3]{\mathcal{M}_{#2}^{#1}\!\left(#3\right)}
\newcommand{\goalTerm}[1]{G^{#1}}
\newcommand{\distTerm}[1]{D^{#1}}
\newcommand{\arcTerm}[1]{A^{#1}}
\newcommand{\visitTerm}[1]{R^{#1}}
\newcommand{\routeCostTerm}[1]{\Psi^{#1}}
\newcommand{\optimalGoalTerm}[1]{G^{#1,\star}}
\newcommand{\feasibleSet}[1]{\mathcal{F}^{#1}}

\newcommand{\episodeSet}{\mathcal{I}}
\newcommand{\episodeHorizon}[1]{T_{#1}}
\newcommand{\evalTargets}[1]{\targetSet_{#1}}
\newcommand{\evalAgents}[1]{\agentSet_{#1}}
\newcommand{\evalCurrentVp}[3]{n_{#1,#2,#3}}
\newcommand{\oracleDetectable}[2]{\mathcal{D}_{#1,#2}}
\newcommand{\detOutput}[4]{\hat y_{#1,#2,#3,#4}}
\newcommand{\verifiedFound}[2]{F_{#1,#2}}

\newcommand{\episodeProgress}[1]{P_{#1}}
\newcommand{\progressScore}{P}
\newcommand{\agentTravelCost}[2]{C_{#1,#2}}
\newcommand{\totalTeamDist}[1]{C_{\Sigma,#1}}
\newcommand{\totalTeamDistOracle}[1]{C_{\Sigma,#1}^{*}}
\newcommand{\makespanTeamDist}[1]{C_{\max,#1}}
\newcommand{\makespanTeamDistOracle}[1]{C_{\max,#1}^{*}}
\newcommand{\teamPPLTotal}{\mathrm{TeamPPL}_{\Sigma}}
\newcommand{\teamPPLMakespan}{\mathrm{TeamPPL}_{\max}}

\newcommand{\oracleGeodesicDist}[3]{d_{#1}^{\mathrm{geo}}(#2,#3)}
\newcommand{\targetOracleDist}[3]{\Delta_{#1,#2}(#3)}
\newcommand{\normTargetOracleDist}[3]{\bar{\Delta}_{#1,#2}(#3)}

\title{
Training-Free Multi-Agent Cooperative Embodied Search from Uncertain Vision-Language Priors
}

\author{Fan~Pu, Keshu~Wu, Jiwan~Jiang, Shiyu~Shen, Zhiwen~Fan,
Zhengzhong~Tu, Jiachen~Li, and~Yang~Zhou%
\thanks{Fan Pu, Keshu Wu, Jiwan Jiang, and Yang Zhou are with the
Department of Civil and Environmental Engineering, Texas A\&M University,
TX 77843.}%
\thanks{Shiyu Shen is with the Department of Civil and Environmental
Engineering, University of Illinois at Urbana-Champaign, IL 61801.}%
\thanks{Zhiwen Fan is with the Department of Electrical and Computer
Engineering, Texas A\&M University, TX 77843.}%
\thanks{Zhengzhong Tu is with the Department of Computer Science and
Engineering, Texas A\&M University, TX 77843.}%
\thanks{Jiachen Li is with the School of Industrial and Systems Engineering
and the School of Mechanical Engineering, Georgia Institute of Technology,
GA 30332.}%
\thanks{Corresponding author: Yang Zhou.}%
}

\begin{document}
\maketitle
\bstctlcite{IEEEexample:BSTcontrol}


\begin{abstract}
Real-world Vision-Language Navigation (VLN) requires agents to act under partial observability, where map structure, semantic cues, and target locations may be incomplete or incorrectly inferred by foundation models. This paper presents a lightweight, training-free, and uncertainty-aware framework for cooperative multi-agent multi-target search in mapless environments. Instead of learning an end-to-end navigation policy, the framework separates high-level semantic reasoning from low-level execution through a scalable graph-optimization backbone. Multimodal Large Language Models (MLLMs) serve only as hypothesis generators, proposing semantic zones, target-location cues, and possible navigable connections. These outputs are treated as uncertain priors and further refined using online observations and graph-context evidence. Cooperative navigation is then formulated as a rolling-horizon multi-agent mixed-integer linear program with a lexicographic objective that first maximizes target-directed reward and then reduces travel distance, uncertain-edge use, and repeated visits. Experiments derived from 100 Matterport3D test cases show that the proposed framework improves target coverage and search efficiency over representative baseline methods, while shared information and joint planning enable more balanced workloads across agents. By operating on an abstract graph without relying on embodiment-specific policy parameters, the framework scales across varying numbers of agents and targets, generalizes to unseen environments, and accommodates different perception modules and low-level controllers without task-specific retraining.
\end{abstract}

\begin{IEEEkeywords}
Multi-agent systems, cooperative search, vision-language navigation, mixed-integer linear programming.
\end{IEEEkeywords}

\section{Introduction}

Robots operating in human environments must make decisions from incomplete and evolving observations. Planning under partial observability and probabilistic robotics provide foundations for reasoning about uncertain states and environment structure \cite{kaelbling1998planning,thrun2005probabilistic}, while mapping and SLAM enable navigation in unknown or partially known spaces \cite{oriolo1998real,dissanayake2001solution,cadena2016past}. Nevertheless, cooperative embodied search remains challenging when multiple robots must locate multiple targets without prior knowledge of the semantic layout of the environment, traversable connections, or target locations.

Vision-Language Navigation (VLN), Object-Goal Navigation (ObjectNav), and their multi-object and open-vocabulary extensions have substantially advanced language-guided navigation and semantic object search \cite{anderson2018vision,fried2018speaker,tan2019learning,ku2020room,krantz2020beyond,savva2019habitat,batra2020objectnav,chaplot2020object,du2023object,chen2023trainingfree,wani2020multion,khanna2024goatbench,majumdar2022zson,gadre2023cows,yokoyama2024hm3d}. Multi-robot task allocation, path planning, and reward-driven routing further provide mechanisms for coordinating agents under shared constraints \cite{gerkey2004formal,korsah2013comprehensive,yu2013structure,stern2019multi,chao1996team,vansteenwegen2011orienteering,zhang2025lamma,gupta2026scale}. 
However, most existing multi-robot formulations coordinate agents over known task representations and sufficiently reliable environment models, with task rewards and graph connectivity largely treated as known during planning.
These assumptions are restrictive in cooperative multi-agent multi-target search, where agents may duplicate one another's exploration, targets compete for limited search effort, and each new observation can alter both target-location beliefs and feasible routes.

Semantic exploration methods address part of this challenge by prioritizing frontiers, informative regions, or locations that are contextually related to a target \cite{yamauchi1997frontier,stachniss2005information,simmons2000coordination,burgard2005coordinated,fox2006distributed,aydemir2011search,kunze2014qualitative,kostavelis2015semantic,gupta2017cognitive}. Recent large language and vision-language models provide an additional source of commonsense knowledge and have been used to propose semantic subgoals, identify promising search regions, construct topological representations, and connect language instructions with robot actions \cite{ahn2022can,shah2023lmnav,driess2023palm,zitkovich2023rt,yu2023l3mvn,zhou2023esc,zhou2024navgpt,wu2024voronav}. Yet these outputs are not verified physical facts. Foundation models may hallucinate objects, semantic regions, or navigable connections \cite{ji2023survey,bai2024hallucination}, making direct use of their predictions as action commands unreliable in partially observed environments.

To address these limitations, this paper employs Multimodal Large Language Models (MLLMs) as structured hypothesis generators for cooperative embodied search. From panoramic observations and accumulated context, the MLLM proposes semantic zones, target-location scores, and possible connections beyond the verified neighborhood. These uncertain hypotheses populate a dynamic semantic topological graph and are updated through physical observations. A rolling-horizon multi-agent Mixed-Integer Linear Program (MILP) first maximizes target-directed reward and then minimizes travel distance, hypothesized-edge uncertainty, and revisits among reward-optimal routes. The abstract graph decouples semantic reasoning from embodiment-specific execution, allowing different perception modules and low-level controllers without retraining. For each current graph, belief state, and planning horizon, the MILP yields a solver-certified optimum, or an explicit optimality gap when terminated early.

The contributions of this paper are threefold:
\begin{itemize}
\item We propose a training-free and uncertainty-aware framework for cooperative multi-agent multi-target search, in which MLLMs generate semantic and topological hypotheses while a graph-based decision layer converts them into coordinated, executable actions.

\item We formulate cooperative search as a rolling-horizon MILP with a two-priority objective that first maximizes uniquely assigned target reward and then minimizes travel distance, hypothesized-edge uncertainty, and revisit costs while preserving the primary optimum.

\item We validate the framework across unseen environments, varying agent and target configurations, and multiple MLLM backbones. The proposed method consistently improves multi-agent target coverage and search efficiency over representative vision-language and decentralized baselines, while the controlled decentralized comparison demonstrates the benefit of shared graph information and joint planning for workload balance and parallel search efficiency.

\end{itemize}

\section{Related Work}

\subsection{Navigation, Mapping, and Embodied Search}

Planning under partial observability and probabilistic robotics provide the foundations for decision-making and state estimation under incomplete information \cite{kaelbling1998planning,thrun2005probabilistic}. In mobile robotics, real-time navigation and map building, SLAM, and modern spatial mapping systems have enabled robots to operate in unknown or partially known environments \cite{oriolo1998real,dissanayake2001solution,cadena2016past}. However, object search requires more than geometric mapping: robots must also reason about semantic regions, likely object locations, and the value of exploring uncertain areas.

VLN and ObjectNav provide two major embodied-navigation formulations related to this work. VLN focuses on following natural-language instructions in visually grounded environments \cite{anderson2018vision,fried2018speaker,tan2019learning,ku2020room,krantz2020beyond,jiang2026navtrust}, whereas ObjectNav asks an agent to search for a specified object category or instance in an unfamiliar scene \cite{savva2019habitat,batra2020objectnav,chaplot2020object,du2023object,chen2023trainingfree}. Recent multi-object, lifelong, and open-vocabulary navigation methods further extend these settings to longer horizons and broader target categories \cite{wani2020multion,khanna2024goatbench,majumdar2022zson,gadre2023cows,yokoyama2024hm3d}. Nevertheless, most of these studies focus on a single agent and do not directly address cooperative multi-agent multi-target search with uncertain map connectivity and dynamically changing target-location estimates.

\subsection{Multi-Agent Planning and Semantic Exploration}

Cooperative search couples task allocation, route construction, and information acquisition. Multi-robot task allocation methods assign robots to tasks under coupled utilities and constraints \cite{gerkey2004formal,korsah2013comprehensive}, while multi-agent path-finding and coverage planning construct coordinated
routes over shared spatial representations \cite{yu2013structure,stern2019multi,lee2026priority}. Team-orienteering formulations optimize reward collection under travel budgets \cite{chao1996team,vansteenwegen2011orienteering}, but typically assume known rewards and graph structure. The difficulty increases under partial observability because one agent's observations can change the value and feasibility of the plans assigned to other agents, consistent with the broader complexity of decentralized planning under uncertainty \cite{bernstein2002complexity}.

Frontier-based and information-gain-based exploration expand unknown environments by prioritizing unexplored or informative regions \cite{yamauchi1997frontier,stachniss2005information}, and coordinated multi-robot exploration reduces redundant coverage by jointly considering information value and travel cost \cite{simmons2000coordination,burgard2005coordinated,fox2006distributed}. Semantic-search methods use object-place relations, semantic maps, memory, and contextual priors to prioritize locations likely to contain a target \cite{aydemir2011search,kunze2014qualitative,kostavelis2015semantic,gupta2017cognitive,chaplot2020object,du2023object}. Vision-language frontier maps further guide exploration by ranking frontiers according to their visual-semantic relevance to the target \cite{yokoyama2024vlfm}. In contrast to these approaches, our work coordinates multiple agents over a graph containing both physically verified transitions and uncertain MLLM-hypothesized connections.

\subsection{Foundation Models for Navigation and Object Search}

Pretrained language and vision-language models have increasingly been incorporated into robotic navigation and decision-making. Existing approaches connect language instructions with feasible actions \cite{ahn2022can}, combine language reasoning with visual navigation \cite{shah2023lmnav}, transfer web-scale vision-language knowledge to embodied control \cite{driess2023palm,zitkovich2023rt}, and use large models to propose semantic subgoals, promising search regions, or topological representations \cite{yu2023l3mvn,zhou2023esc,zhou2024navgpt,wu2024voronav}. These methods demonstrate that foundation models can provide useful high-level guidance without environment-specific training.

However, hallucination remains a persistent limitation of both language and multimodal foundation models \cite{ji2023survey,bai2024hallucination}. In embodied settings, hallucinations can appear as incorrect target-location predictions, non-existent semantic regions, or infeasible spatial connections. The proposed framework therefore differs from methods that directly translate foundation-model outputs into actions. It treats MLLM outputs as uncertain graph hypotheses, revises them using physical observations and graph-context evidence, and delegates coordinated action selection to an explicit optimization model with great generalizability.

\section{Problem Formulation}\label{sec:problem}
\begin{figure*}[t]
    \centering
    \begin{minipage}{1\textwidth}
        \centering
        \includegraphics[width=\linewidth]{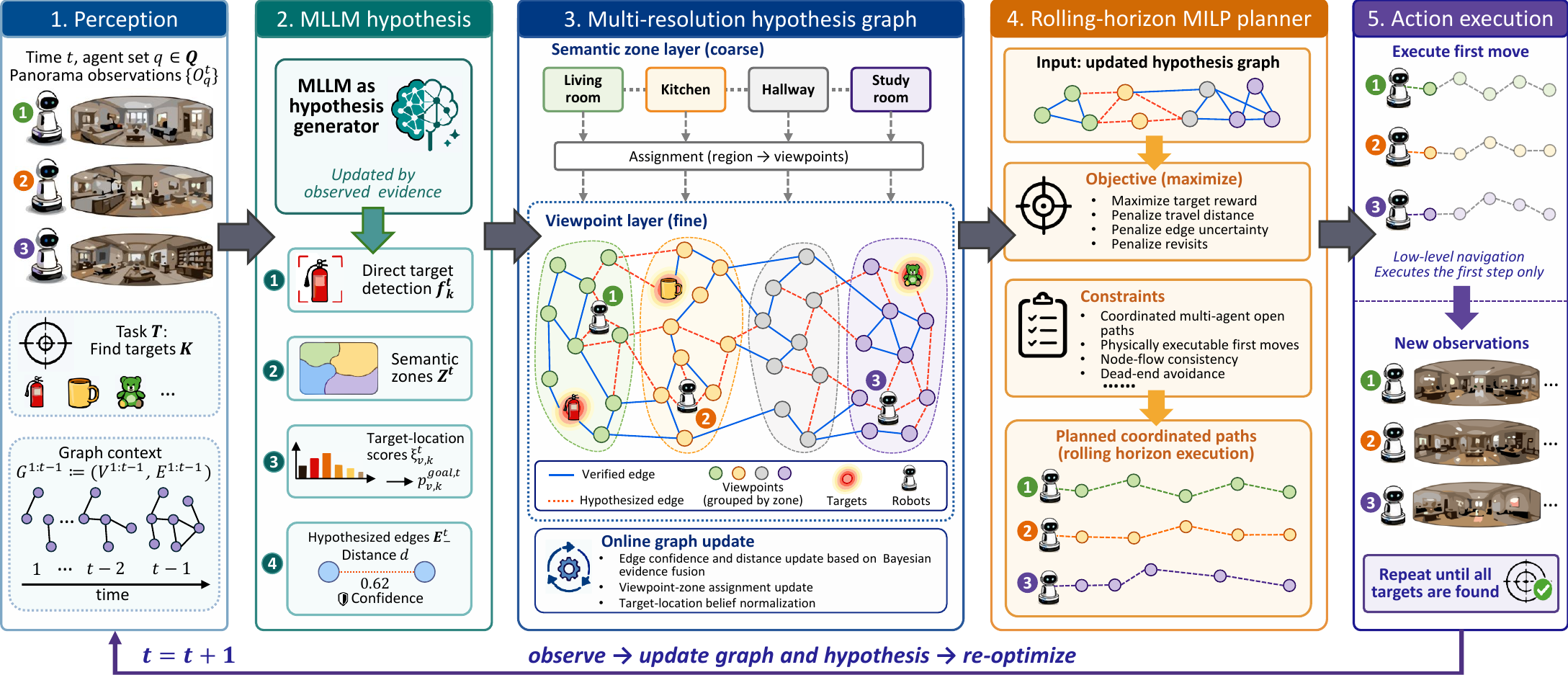}
        \caption{Overall architecture of the proposed framework.}
        \label{fig:fram_illus}
    \end{minipage}
\end{figure*}

We consider cooperative embodied search in partially observed, mapless environments, where a team of robots $\agentSet$ must locate multiple targets $\targetSet$ without prior knowledge of the environment topology, semantic regions, traversable connections, or target locations. As illustrated in \autoref{fig:fram_illus}, the proposed framework follows a closed-loop hypothesize--verify--optimize--execute process. At each decision step, the robots acquire local panoramic observations and feasible one-hop actions; an MLLM then uses these observations, the task descriptions, and the accumulated graph context to propose uncertain semantic and topological hypotheses, including target-location scores and possible unobserved connections. These hypotheses are then incorporated into a dynamic graph that distinguishes physically verified edges from ungrounded MLLM-proposed edges. A rolling-horizon MILP selects coordinated routes for the team, of which only the first grounded transition is executed before new observations are collected, the graph is updated, and the problem is solved again.

\subsection{Partial Observations and Graph Context}
\begin{figure}[t]
    \centering
    \includegraphics[width=0.7\linewidth]{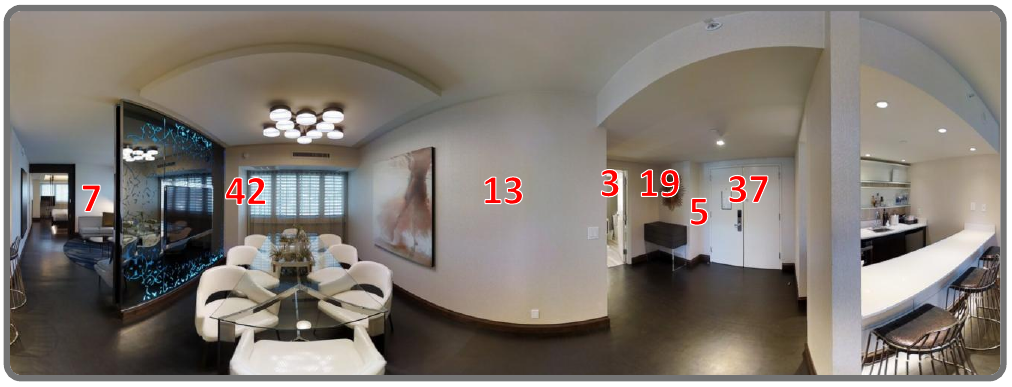}
    \caption{Panoramic observation at an agent's current viewpoint. The red numbers denote the IDs of neighboring viewpoints returned by the navigation system and constitute the agent's feasible one-hop action space.}
    \label{fig:observe_images}
\end{figure}

At time $t$, the framework maintains a graph $\graph{1:t-1}$ containing the geometric, semantic, and hypothesized information accumulated through time $t-1$. For any graph component, superscript $t$ denotes information generated or observed at time $t$ before integration, whereas $1:t$ denotes the retained cumulative information.
%
Each agent $q$, located at viewpoint $\orCurrentVp{q}$, receives $\observ{t}{q}=(\textTargetSet,\rgbImage{\orCurrentVp{q}},\actionSpace{\orCurrentVp{q}})$, where $\textTargetSet$ contains the target descriptions, $\rgbImage{\orCurrentVp{q}}$ is the horizontal panorama observed at the current viewpoint, and $\actionSpace{\orCurrentVp{q}}$ contains the neighboring viewpoints that are reachable in one step together with their travel distances. \autoref{fig:observe_images} shows an example panorama. The red labels identify the neighboring viewpoint IDs returned by the navigation system; in this example, they form the feasible one-hop action set available from the current viewpoint.
These local observations and feasible actions provide the current perceptual input for graph generation, while $\graph{1:t-1}$ supplies the accumulated context from previous steps. The MLLM then combines $\{\observ{t}{q}:q\in\agentSet\}$ with $\graph{1:t-1}$ to generate updated environment hypotheses and form $\graph{1:t}$. The planner then executes only the first selected action before obtaining new observations and replanning at time $t+1$. Thus, the graph is progressively grounded and refined during exploration.

\subsection{Multi-Resolution Hypothesized Representation}
From $\{\observ{t}{q}:q\in\agentSet\}$, the MLLM proposes semantic zones
$\regionZones{t}:=\{\regionCurrent{t}{q}:q\in\agentSet\}\cup\regionDist{t}$,
where $\regionCurrent{t}{q}$ describes the area currently occupied by agent $q$, and $\regionDist{t}$ contains directly observed distal areas. These zones provide coarse semantic context for viewpoint assignment, target-location scoring, and graph consistency.
The viewpoint layer contains the current and reachable poses: $\viewpoints{t}
:=\bigcup_{q\in\agentSet}\actionSpace{\orCurrentVp{q}}\cup\{\orCurrentVp{q}:q\in\agentSet\}.$
The MLLM assigns each viewpoint to one semantic zone, and $\vpAssigns{1:t}{n}$ denotes the viewpoints assigned to zone $n$. Thus, semantic zones provide high-level context, while viewpoints and their connections support navigation and action execution.


\subsection{Foundation-Model Graph Hypotheses}
At time $t$, only targets that have not yet been directly detected remain active. We define the binary indicator $\targetFound{t}{k}$, which equals to 1 if target $k$ has been found as of time $t$. Therefore, we define $\actTargets{t}:=\left\{k\in\targetSet:\targetFound{t}{k}=0\right\}$.
For any found target $k\notin\actTargets{t}$, target-location probabilities and raw target scores are removed from all nodes and are no longer included in graph-generation prompts.

\paragraph{\textbf{Node-level hypothesis}}
For each active target $k\in\actTargets{t}$, the MLLM assigns nonnegative relative scores to eligible viewpoints based on the current observations and graph context; these scores are normalized into target-location probabilities used by the optimizer.
Let $\currentVPSet{t}:=\{\orCurrentVp{q}:q\in\agentSet\}$ denote the set of current viewpoints. The eligible viewpoint set is defined as
\begin{equation}
\eligibleVPs{t}
:=
\left\{
v\in\viewpoints{1:t}:
v\notin\currentVPSet{t},~
\nodeVisitTimes{t}{v}=0
\right\},
\label{eq:eligible_vps}
\end{equation}
where $\nodeVisitTimes{t}{v}$ indicates the times that viewpoint $v$ has been visited as of time $t$.
For each $v\in\eligibleVPs{t}$ and $k\in\actTargets{t}$, let $\rawVPGoalScore{t}{v}{k}\ge 0$ denote the raw target-location score. When the total score for target $k$ is positive, the normalized target-location probability is given by $\targetProb{t}{v}{k} = \rawVPGoalScore{t}{v}{k}/\sum_{u\in\eligibleVPs{t}}\rawVPGoalScore{t}{u}{k}$.

We say that a viewpoint $v$ is grounded if it has been visited, that is, $\nodeVisitTimes{t}{v}>0$. For a grounded viewpoint $v$, an active target that is not directly detected at $v$ is treated as absent from that viewpoint for planning purposes. Therefore, for any $v\in\viewpoints{1:t}\setminus\eligibleVPs{t},~
k\in\actTargets{t}$, we fix $\targetProb{t}{v}{k}=0$.
For any found target $k\notin\actTargets{t}$, its raw scores and target-location probabilities are removed from the graph state.

\paragraph{\textbf{Edge-level hypothesis}}
We define edges to describe possible navigable relations between feasible agent poses. At time slot $t$, let $\VPEdges{t}:=\VPEdgesGrounded{t}\cup\VPEdgesUngrounded{t}$ be the edges generated during time slot $t$, which consists of two disjoint subsets $\VPEdgesGrounded{t}$ and $\VPEdgesUngrounded{t}$.
The set $\VPEdgesGrounded{t}$ contains local transitions returned by the navigation system from the agents' current viewpoints to their visible neighboring viewpoints. These edges are physically verified and have measured travel distances. In contrast, $\VPEdgesUngrounded{t}$ are MLLM hypothesized edges proposed from the current observations and the accumulated graph context, which represent plausible navigable connections between viewpoint nodes that have not yet been physically visited. When edge hypotheses are available, ungrounded edges support look-ahead planning beyond the physically verified local neighborhood and remain uncertain until their connectivity is verified as exploration reaches the corresponding endpoints.

For each physically observed edge $\edgeWrap{i}{j}\in\VPEdgesGrounded{t}$, its existence is verified and the travel distance $\travelDist{t}{\edgeWrap{i}{j}}$ is directly measured. For each hypothesized edge $\edgeWrap{i}{j}\in\VPEdgesUngrounded{t}$, let $\edgeExistLatent{\edgeWrap{i}{j}}\in{0,1}$ indicate whether the proposed connection is truly navigable. Before physical verification, the MLLM assigns an initial conditional edge-existence probability based on the agents' current observations, the accumulated graph context through time $t-1$, and the target descriptions, as
{\small
\begin{equation}
\begin{aligned}
    \MLLMcondEdgeExistProb{t}{\edgeWrap{i}{j}}
    :=
    \mathbb{P}~\bigl(
    \edgeExistLatent{\edgeWrap{i}{j}}=1
    ~|~
    &\{\observ{t}{q}:q\in\agentSet\},~
    \graph{1:t-1},~
    \textTargetSet
    \bigr).
\end{aligned}
\label{eq:init_cond_edge_exist_vv}
\end{equation}
}

Moreover, the MLLM also provides an initial distance estimate $\MLLMEdgeDist{t}{\edgeWrap{i}{j}}$ for each newly hypothesized edge $\edgeWrap{i}{j}\in\VPEdgesUngrounded{t}$. This estimate represents the expected travel distance if the hypothesized connection exists. Its initial uncertainty is represented by the variance parameter $\MLLMdistVarVV{t}{\edgeWrap{i}{j}}$.
For each retained edge $\edgeWrap{i}{j}$, $\condEdgeExistProb{t}{\edgeWrap{i}{j}}$ denotes its conditional existence probability at time $t$. For each retained ungrounded edge, $\travelDist{t}{\edgeWrap{i}{j}}$ and $\travelDistVar{t}{\edgeWrap{i}{j}}$ denote its current distance estimate and variance.

\subsection{Dynamic Graph Update}
\paragraph{\textbf{Node update}}
At time slot $t$, the MLLM proposes a set of semantic-zone hypotheses $\regionZones{t}$ from the current observations, while the onboard navigation system provides the feasible viewpoint set $\viewpoints{t}$. These newly identified semantic zones and viewpoints are incorporated into the accumulated graph only when they have not appeared in previous time slots. Thus, the cumulative node sets are updated as
\begin{equation}
\regionZones{1:t} =\regionZones{1:t-1}\cup\regionZones{t}, \qquad \viewpoints{1:t}=\viewpoints{1:t-1}\cup\viewpoints{t}.
\end{equation}

\paragraph{\textbf{Viewpoint visit-status update}}
Once each agent executes an action and reaches a new pose, its corresponding current viewpoint node becomes physically grounded. We require different agents to occupy distinct current viewpoint nodes at each time slot, that is, $\orCurrentVp{q}\neq\orCurrentVp{q'}$ for all $q,q'\in\agentSet$ with $q\neq q'$.
Therefore, for the current viewpoint node $\orCurrentVp{q}$ of each agent $q\in\agentSet$, we mark the node as grounded and update its visit count as $\nodeVisitTimes{t}{\orCurrentVp{q}}=\nodeVisitTimes{t-1}{\orCurrentVp{q}}+1$, where $\nodeVisitTimes{t-1}{v}=0$ if viewpoint node $v$ is newly introduced at time slot $t$.

\paragraph{\textbf{Viewpoint assignment update}}
Let $\zoneOfVP{t}{v}\in\regionZones{1:t}$ denote the unique semantic zone assigned to viewpoint node $v\in\viewpoints{1:t}$ at time slot $t$ and $\vpAssigns{1:t}{n}$ is the viewpoint-assignment set:
\begin{equation}\label{eq:vpAssigns}
\vpAssigns{1:t}{n}:=\{v\in\viewpoints{1:t}:~\zoneOfVP{t}{v}=n\},
\quad
\forall n\in\regionZones{1:t}.
\end{equation}

For a newly introduced viewpoint node $v\in\viewpoints{t}\setminus\viewpoints{1:t-1}$, the assignment $\zoneOfVP{t}{v}$ is initialized by the MLLM based on $\{\observ{t}{q}:q\in\agentSet\}$ and $\graph{1:t-1}$, so that every new viewpoint is assigned to one semantic zone upon entering the graph. For a viewpoint node $v\in\viewpoints{1:t-1}$ that already exists in the graph, its previous assignment is preserved, that is, $\zoneOfVP{t}{v}=\zoneOfVP{t-1}{v}$, unless $v$ becomes grounded at the current time slot. In that case, the assignment is overridden by the current semantic zone, that is,
\begin{equation}
\zoneOfVP{t}{\orCurrentVp{q}}
=
\regionCurrent{t}{q},
\quad
\forall q\in\agentSet:
\nodeVisitTimes{t-1}{\orCurrentVp{q}}=0,~
\nodeVisitTimes{t}{\orCurrentVp{q}}=1.
\end{equation}

This override is imposed because a viewpoint grounded for the first time at time slot $t$ corresponds to the agent's current location and should therefore be assigned to the proximal semantic zone inferred from the current observation.

\paragraph{\textbf{Edge update}}
Let $\VPEdges{1:t}$ denote the retained cumulative edge set after the update at time $t$; thus, $\VPEdges{1:t}:=\VPEdges{1:t-1}\cup\VPEdges{t}$. We partition $\VPEdges{1:t}$ into grounded edges $\VPEdgesGrounded{1:t}$ and ungrounded edges $\VPEdgesUngrounded{1:t}$. An edge is grounded if at least one of its endpoint viewpoints is grounded, so that its existence is verified by the navigation system at that viewpoint. Accordingly, a grounded edge represents a physically verified executable movement, whereas an ungrounded edge represents a hypothesized connection that has not yet been physically verified.

Recall that $\actionSpace{\orCurrentVp{q}}$ denotes the action space of the current viewpoint of agent $q$. The edges $\obsVVEdges{t}:=\cup_{q\in\agentSet,j\in\actionSpace{\orCurrentVp{q}}} \edgeWrap{\orCurrentVp{q}}{j}$ then represent the set of physically observed transitions returned by the navigation system at time slot $t$, connecting each current viewpoint to its visible neighboring viewpoints with measured travel distances $\travelDist{t}{\edgeWrap{i}{j}}$. We update the grounded edge set as $\VPEdgesGrounded{1:t}=\VPEdgesGrounded{1:t-1}\cup\obsVVEdges{t}$.
By definition, every edge $\edgeWrap{i}{j}$ in $\obsVVEdges{t}$ is grounded, and therefore $\condEdgeExistProb{t}{\edgeWrap{i}{j}}=1$. 
The ungrounded edge set contains only hypothesized edges whose endpoints remain ungrounded; hence, $\VPEdgesUngrounded{1:t}
=\left(
\VPEdgesUngrounded{1:t-1}
\cup
\VPEdgesUngrounded{t}
\right)
\setminus
\obsVVEdges{t}.$

\subsection{Uncertain Hypothesis Update}
Based on the updated graph, we revise the remaining uncertain hypotheses using the current observations and accumulated graph context. Target-location probabilities are updated from MLLM predictions over eligible viewpoints. The remaining uncertainty is primarily associated with hypothesized edges $\VPEdgesUngrounded{1:t}$, whose existence likelihoods and distance estimates are refined as additional evidence becomes available.

\paragraph{\textbf{Edge distance hypothesis update}}
The distance-update rules apply only to ungrounded edges $\VPEdgesUngrounded{1:t}$ based on Bayesian Gaussian fusion.
For each ungrounded edge $\edgeWrap{i}{j}\in\VPEdgesUngrounded{1:t}$, let $\edgeDistLatent{\edgeWrap{i}{j}}$ denote its latent true travel distance, conditioned on the event that edge $\edgeWrap{i}{j}$ exists. Before incorporating the current observations $\{\observ{t}{q}:q\in\agentSet\}$, let $\edgeDistPriorMean{t}{\edgeWrap{i}{j}}$ and $\edgeDistPriorVar{t}{\edgeWrap{i}{j}}$ denote the prior mean and prior variance of $\edgeDistLatent{\edgeWrap{i}{j}}$, respectively. For edges already present in $\graph{1:t-1}$, the prior mean and variance are inherited from the previous posterior estimates, that is, $\edgeDistPriorMean{t}{\edgeWrap{i}{j}}=\travelDist{t-1}{\edgeWrap{i}{j}}$ and $\edgeDistPriorVar{t}{\edgeWrap{i}{j}}=\travelDistVar{t-1}{\edgeWrap{i}{j}}$.
For newly introduced ungrounded edges in $\VPEdgesUngrounded{t}$ at time $t$, the prior mean and prior variance of the edge distance are initialized by the MLLM-provided estimate, that is, $\edgeDistPriorMean{t}{\edgeWrap{i}{j}}=\MLLMEdgeDist{t}{\edgeWrap{i}{j}}$ and $\edgeDistPriorVar{t}{\edgeWrap{i}{j}}=\MLLMdistVarVV{t}{\edgeWrap{i}{j}}$.

Because the initial distance hypothesis of a newly introduced edge is generated from the current observations, it is not fused again with evidence from the same time slot. Gaussian fusion is applied only to ungrounded edges that were already present in $\VPEdgesUngrounded{1:t-1}$.
Conditioned on the edge-existence hypothesis, we assume the prior model
\begin{equation}
\edgeDistLatent{\edgeWrap{i}{j}}\mid \big(\edgeExistLatent{\edgeWrap{i}{j}}=1,\graph{1:t-1},\textTargetSet\big) \sim \mathcal{N}\left( \edgeDistPriorMean{t}{\edgeWrap{i}{j}},\edgeDistPriorVar{t}{\edgeWrap{i}{j}}\right).
\label{eq:edge_distance_prior}
\end{equation}

At time slot $t$, the current observation and graph context provide a cue-based distance estimate $\cueEdgeDist{t}{\edgeWrap{i}{j}}$ for edge $\edgeWrap{i}{j}$. We model this cue as
\begin{equation}
\begin{aligned}
    \cueEdgeDist{t}{\edgeWrap{i}{j}} \mid \big(\edgeDistLatent{\edgeWrap{i}{j}},\edgeExistLatent{\edgeWrap{i}{j}}=1,\{\observ{t}{q}:q\in\agentSet\},\graph{1:t-1},\textTargetSet\big)\\
    \sim\mathcal{N}\left(\edgeDistLatent{\edgeWrap{i}{j}},\cueEdgeDistVar{t}{\edgeWrap{i}{j}}\right),
\end{aligned}
\label{eq:edge_distance_cue_model}
\end{equation}
where $\cueEdgeDistVar{t}{\edgeWrap{i}{j}}$ is the cue variance, which quantifies the uncertainty of the current observation-derived estimate. When an informative cue is available, the Gaussian-Gaussian conjugate update yields the following Gaussian posterior:
\begin{equation}
\begin{aligned}
    \edgeDistLatent{\edgeWrap{i}{j}}\mid \big(\edgeExistLatent{\edgeWrap{i}{j}}=1, \{\observ{t}{q}:q\in\agentSet\},\graph{1:t-1},\textTargetSet \big)\\
    \sim\mathcal{N}\left(\travelDist{t}{\edgeWrap{i}{j}},\travelDistVar{t}{\edgeWrap{i}{j}}\right),
\end{aligned}
\label{eq:edge_distance_posterior}
\end{equation}
with $\travelDistVar{t}{\edgeWrap{i}{j}} =$
{\small
\begin{equation}
\begin{cases} \displaystyle \left( \frac{1}{\edgeDistPriorVar{t}{\edgeWrap{i}{j}}} + \frac{1}{\cueEdgeDistVar{t}{\edgeWrap{i}{j}}} \right)^{-1}, &
\edgeWrap{i}{j}\in\VPEdgesUngrounded{1:t-1} \\
\MLLMdistVarVV{t}{\edgeWrap{i}{j}}, & \edgeWrap{i}{j}\in \VPEdgesUngrounded{t} \setminus \VPEdgesUngrounded{1:t-1},
\end{cases} \label{eq:edge_distance_post_var}
\end{equation}
}

For $\travelDist{t}{\edgeWrap{i}{j}}$, if $\edgeWrap{i}{j}\in\VPEdgesUngrounded{1:t-1}$, then
\begin{equation}
\travelDist{t}{\edgeWrap{i}{j}} = \travelDistVar{t}{\edgeWrap{i}{j}} \left( \frac{ \edgeDistPriorMean{t}{\edgeWrap{i}{j}} }{ \edgeDistPriorVar{t}{\edgeWrap{i}{j}} } + \frac{ \cueEdgeDist{t}{\edgeWrap{i}{j}} }{ \cueEdgeDistVar{t}{\edgeWrap{i}{j}} } \right),
\label{eq:edge_distance_post_mean1}
\end{equation}
and otherwise if $\edgeWrap{i}{j} \in \VPEdgesUngrounded{t}\setminus\VPEdgesUngrounded{1:t-1}$, then $\travelDist{t}{\edgeWrap{i}{j}} =\MLLMEdgeDist{t}{\edgeWrap{i}{j}}$.

For an uncertain edge already present in $\graph{1:t-1}$, the updated distance $\travelDist{t}{\edgeWrap{i}{j}}$ is obtained through a precision-weighted fusion of the historical estimate and the current cue-based estimate, whereas for a newly introduced edge, the current MLLM-provided distance and variance are retained.
The remaining task is therefore to define the cue mean $\cueEdgeDist{t}{\edgeWrap{i}{j}}$ and cue variance $\cueEdgeDistVar{t}{\edgeWrap{i}{j}}$ for ungrounded edges.

Given that, for $\edgeWrap{i}{j}\in\VPEdgesUngrounded{1:t}$, one informative current cue is the empirical distance pattern of grounded edges, that is, the empirical mean $\empMeanEdgeDist{t}$ and variance $\empVarEdgeDist{t}$ of the distances of existing grounded edge, as defined below:
\begin{gather}
    \empMeanEdgeDist{t}:=\frac{1}{|\VPEdgesGrounded{1:t}|}\sum_{\edgeWrap{i}{j}\in\VPEdgesGrounded{1:t}}\travelDist{t}{\edgeWrap{i}{j}},\\
    \empVarEdgeDist{t}:=\frac{1}{|\VPEdgesGrounded{1:t}|}\sum_{\edgeWrap{i}{j}\in\VPEdgesGrounded{1:t}}\bigl(\travelDist{t}{\edgeWrap{i}{j}} - \empMeanEdgeDist{t}\bigr)^2+\smallNumber,
\end{gather}
where $\smallNumber>0$ is a small constant for numerical stability.

Another informative current cue is the semantic support that nodes $i$ and $j$ belong to a common semantic zone. For this, let $\vpAssignIndicator{t}{\edgeWrap{i}{j}}$ be the binary indicator $\vpAssignIndicator{t}{\edgeWrap{i}{j}}$ which equals 1 if viewpoint nodes $i$ and $j$ are assigned to the same semantic zone, and equals 0 otherwise.
Accordingly, when $\VPEdgesGrounded{1:t}\neq\emptyset$, the empirical distance statistics provide the cue mean and cue variance for each $\edgeWrap{i}{j}\in\VPEdgesUngrounded{1:t}$:
\begin{equation} \cueEdgeDist{t}{\edgeWrap{i}{j}} := \empMeanEdgeDist{t}, \qquad \cueEdgeDistVar{t}{\edgeWrap{i}{j}} := \frac{ \empVarEdgeDist{t} }{ 1+\VVedgeCueFactor \vpAssignIndicator{t}{\edgeWrap{i}{j}}},
\label{eq:edge_distance_vv_cue_mean}
\end{equation}
where $\VVedgeCueFactor>0$ is a fixed parameter. Thus, the distance cue encourages the hypothesized ungrounded edge distance to remain close to the empirical local-transition scale.

Because the plausibility of a hypothesized connection also depends on whether its estimated distance is consistent with observed local transitions, the distance information above is further used as evidence when updating edge-existence probabilities.

\paragraph{\textbf{Edge existence hypothesis update}}
For each grounded edge $\edgeWrap{i}{j}\in\VPEdgesGrounded{1:t}$, its existence is physically verified, and hence $\condEdgeExistProb{t}{\edgeWrap{i}{j}}=1$. For each ungrounded edge $\edgeWrap{i}{j}\in\VPEdgesUngrounded{1:t}$, let $\edgeExistLatent{\edgeWrap{i}{j}}\in\{0,1\}$ denote its latent edge-existence indicator, where $\edgeExistLatent{\edgeWrap{i}{j}}=1$ indicates that the hypothesized connection is navigable. Because all endpoint viewpoints are supplied by the navigation system and are therefore known to exist, uncertainty is associated only with the connection between them. Accordingly, $\condEdgeExistProb{t}{\edgeWrap{i}{j}}$ directly represents the existence probability of the hypothesized edge.

Let $\priorCondEdgeExistProb{t}{\edgeWrap{i}{j}}$ denote the prior conditional existence probability before applying the current evidence-based reweighting. For an ungrounded edge already present before time slot $t$, this prior is inherited from the previous posterior. For a newly introduced ungrounded edge, it is initialized by the MLLM-provided conditional edge-existence probability. Thus,
\begin{equation}
\priorCondEdgeExistProb{t}{\edgeWrap{i}{j}}
=
\begin{cases}
\condEdgeExistProb{t-1}{\edgeWrap{i}{j}},
&
\edgeWrap{i}{j}
\in
\VPEdgesUngrounded{1:t-1},
\\
\MLLMcondEdgeExistProb{t}{\edgeWrap{i}{j}},
&
\edgeWrap{i}{j}
\in
\VPEdgesUngrounded{1:t}
\setminus
\VPEdgesUngrounded{1:t-1}.
\end{cases}
\label{eq:conditional_edge_exist_prior}
\end{equation}

A newly introduced ungrounded edge $\edgeWrap{i}{j}\in\VPEdgesUngrounded{1:t}\setminus\VPEdgesUngrounded{1:t-1}$ retains its MLLM-provided existence confidence because this confidence has already been inferred from the current observations and graph context. Therefore, $\condEdgeExistProb{t}{\edgeWrap{i}{j}}=\priorCondEdgeExistProb{t}{\edgeWrap{i}{j}}$.
For an ungrounded edge $\edgeWrap{i}{j}\in\VPEdgesUngrounded{1:t-1}$, the existence probability is reweighted using two graph-based cues: whether its endpoint viewpoints are assigned to the same semantic zone, and whether its prior distance estimate is consistent with the empirical distance pattern of grounded edges. The prior distance estimate $\edgeDistPriorMean{t}{\edgeWrap{i}{j}}$ is used in this update so that the current distance-fusion result does not enter the edge-existence update again through the posterior distance.

For each previously existing ungrounded edge, define the combined evidence score as
{\small
\begin{equation}
    \edgeCompScore{t}{\edgeWrap{i}{j}}:=\log\left(\frac{\edgeExistLikelihoodFactor}{1-\edgeExistLikelihoodFactor}\right)\left(2\vpAssignIndicator{t}{\edgeWrap{i}{j}}-1\right)-\frac{\left(\edgeDistPriorMean{t}{\edgeWrap{i}{j}}-\empMeanEdgeDist{t}\right)^2}{2\empVarEdgeDist{t}},
\label{eq:edge_vv_comp_score}
\end{equation}
}
where $1/2<\edgeExistLikelihoodFactor<1$ is a fixed parameter. The first term provides positive evidence when the two viewpoints are assigned to the same semantic zone and negative evidence otherwise. The second term penalizes the edge-existence hypothesis if the prior distance estimate deviates from the empirical distance pattern of physically verified local transitions.

Using this score, we define the positive evidence weights under the edge-existence and edge-nonexistence hypotheses as
{\small
\begin{equation}
\edgeExistLikelihood{t}{\edgeWrap{i}{j}}
:=
\exp\left(
\frac{1}{2}
\edgeCompScore{t}{\edgeWrap{i}{j}}
\right),
\quad
\edgeNonExistLikelihood{t}{\edgeWrap{i}{j}}
:=
\exp\left(
-\frac{1}{2}
\edgeCompScore{t}{\edgeWrap{i}{j}}
\right),
\label{eq:edge_exist_likelihood_vv}
\end{equation}
}
which are likelihood surrogates that remain strictly positive. Then, for each previously existing ungrounded edge $\edgeWrap{i}{j}\in\VPEdgesUngrounded{1:t-1}$, the posterior conditional existence probability is
\begin{equation}
\condEdgeExistProb{t}{\edgeWrap{i}{j}}
=
\frac{
\edgeExistLikelihood{t}{\edgeWrap{i}{j}}\,
\priorCondEdgeExistProb{t}{\edgeWrap{i}{j}}
}{
\edgeExistLikelihood{t}{\edgeWrap{i}{j}}\,
\priorCondEdgeExistProb{t}{\edgeWrap{i}{j}}
+
\edgeNonExistLikelihood{t}{\edgeWrap{i}{j}}\,
\left(
1-\priorCondEdgeExistProb{t}{\edgeWrap{i}{j}}
\right)
}.
\label{eq:edge_exist_bayes_compact}
\end{equation}

\paragraph{\textbf{Target-location hypothesis update}}
For each eligible viewpoint $v\in\eligibleVPs{t}$ and active target $k\in\actTargets{t}$, let $\rawVPGoalScore{t}{v}{k}\ge 0$ denote the raw target-location score. This score is not treated as a calibrated probability. It is a relative target-location score used to compare eligible viewpoint nodes for the same active target. Existing eligible viewpoints retain their previous raw scores unless the MLLM updates them using the current observations and graph context. Newly eligible viewpoints that do not have prior raw scores must receive raw scores for all active targets.
For each active target $k\in\actTargets{t}$, the raw scores are normalized across eligible viewpoint nodes:
\begin{equation}
\targetProb{t}{v}{k}
=
\frac{\rawVPGoalScore{t}{v}{k}}
{\sum_{u\in\eligibleVPs{t}}\rawVPGoalScore{t}{u}{k}},
\qquad
\forall v\in\eligibleVPs{t},~k\in\actTargets{t},
\label{eq:viewpoint_target_score_norm}
\end{equation}
which ensures that $0\le\targetProb{t}{v}{k}\le 1$.

For detection-fixed viewpoint nodes, target-location probabilities and raw scores for active targets are set to zero:
\begin{equation}
\targetProb{t}{v}{k}=0,
\quad
\rawVPGoalScore{t}{v}{k}=0,
\quad
\forall k\in\actTargets{t},~v\in\viewpoints{1:t}\setminus\eligibleVPs{t}.
\label{eq:detection_fixed_viewpoint_zero}
\end{equation}

Semantic regions do not carry target-location scores and they are retained only as semantic grouping nodes for viewpoint assignment and graph-context construction.
For any found target $k\notin\actTargets{t}$, its raw scores and target-location probabilities are removed from all retained graph nodes. Therefore, subsequent graph generation and optimization consider only the remaining active targets.

\subsection{Optimization Model Design}
\label{sec:opt_model}
We formulate cooperative multi-agent search as a lexicographic Mixed-Integer Linear Program (MILP) that jointly determines which agent should pursue each active target at candidate viewpoints and constructs the corresponding agent routes over the current hypothesis graph. The routing component follows a network-flow formulation, where binary arc-selection variables and flow-conservation constraints construct an open path for each agent \cite{wolsey2014integer}. A lexicographic objective enforces strict priority among optimization criteria, ensuring that the primary target-search objective is optimized before secondary routing costs are considered \cite{ehrgott2005multicriteria}.

We consider multiple agents $\agentSet$ searching for multiple target objects $\targetSet$ in a shared hypothesis graph $\graph{1:t}$. All agents share the same graph state, while each agent constructs its own open path from its current viewpoint. The optimization is solved in a rolling-horizon manner: the model plans a full path over the current hypothesis graph, but only the first physically executable transition is implemented before the graph is updated and the problem is solved again.

In the optimization model, paths are constructed only over viewpoint nodes because executable navigation is represented by edge connectivity. Semantic regions are retained in the hypothesis graph for viewpoint assignment, semantic context, and target-location reasoning, but they are not routing nodes. Therefore, we define $\orNodes{t}:=\viewpoints{1:t}$ as the routing-node set. In particular, each agent's current viewpoint $\orCurrentVp{q}\in\orNodes{t}$ serves as the fixed origin of its planned route.

Since the hypothesis graph stores edges as undirected edges, each edge $\edgeWrap{i}{j}\in\VPEdges{1:t}$ is converted into two directed arcs $\orEdgeWrap{i}{j}$ and $\orEdgeWrap{j}{i}$ in the optimization model. Thus, the directed routing-arc set is $\routingArcs{t}:=\left\{\orEdgeWrap{i}{j},\orEdgeWrap{j}{i}: \edgeWrap{i}{j}\in\VPEdges{1:t}\right\}$.
Each directed arc $\orEdgeWrap{i}{j}\in\routingArcs{t}$ inherits the travel distance $\travelDist{t}{\edgeWrap{i}{j}}$ and conditional edge-existence probability $\condEdgeExistProb{t}{\edgeWrap{i}{j}}$ of its underlying undirected edge.

For each agent $q\in\agentSet$, we define $\candidateNodes{t}{q}:=\orNodes{t}\setminus\{\orCurrentVp{q}\}$ as the set of non-origin routing nodes whose inclusion in the planned route is determined by the optimization model. The current viewpoint $\orCurrentVp{q}$ is excluded from $\candidateNodes{t}{q}$ because it is already fixed as the route origin. The reward-endpoint set $\rewardVPs{t}{q}\subseteq\candidateNodes{t}{q}$ consists of the reachable, unvisited viewpoints outside $\currentVPSet{t}$ that remain after the dead-end exclusions introduced below.

\paragraph{\textbf{Decision Variables}}
For all $\orEdgeWrap{i}{j}\in\routingArcs{t}$ and $q\in\agentSet$, we define
\begin{equation*}
\arcSelectVar{t}{i,j}{q}
=
\begin{cases}
1, & \text{if agent }q\text{ uses directed arc }\orEdgeWrap{i}{j},\\
0, & \text{otherwise}.
\end{cases}
\end{equation*}

For all $n\in\candidateNodes{t}{q}$ and $q\in\agentSet$, we introduce the auxiliary node-selection variable
\begin{equation*}
\multiAgentNodeSelectVar{t}{n}{q}
=
\begin{cases}
1, & \text{if agent } q \text{ includes node } n \text{ in its planned path},\\
0, & \text{otherwise},
\end{cases}
\end{equation*}
which is introduced to simplify the model representation and does not introduce an independent decision, since its value is determined by the arc-selection variables $\arcSelectVar{t}{i,j}{q}$ through equality constraints introduced below.

For each node $n\in\orNodes{t}$ and agent $q\in\agentSet$, let $\multiAgentSubtourVar{t}{n}{q}$ be a continuous ordering variable used for subtour elimination. In addition, for each target $k\in\targetSet$, reward endpoint $v\in\rewardVPs{t}{q}$, and agent $q\in\agentSet$, define
{\small
\begin{equation*}
\targetAssignVar{t}{k}{v}{q}
=
\begin{cases}
1, & \text{if target }k\text{ is assigned to endpoint }v\text{ for agent }q,\\
0, & \text{otherwise},
\end{cases}
\end{equation*}
}

The viewpoint index $v$ is retained because target reward is viewpoint-dependent. Allowing each target to contribute through at most one agent--viewpoint pair also prevents diffuse or noisy target-location scores from accumulating across multiple viewpoints.
We further let $x$, $y$, $u$, and $a$ denote the complete collections of arc-selection, node-selection, ordering, and target-assignment
variables defined above, respectively.

\paragraph{\textbf{Lexicographic Objective}}
Let $\feasibleSet{t}$ denote the feasible set defined by the variable domains and constraints below. The primary objective maximizes the processed viewpoint-level target probabilities:
\begin{equation}
\goalTerm{t}
:=
\sum_{q\in\agentSet}
\sum_{v\in\rewardVPs{t}{q}}
\sum_{k\in\actTargets{t}}
\targetProb{t}{v}{k}
\targetAssignVar{t}{k}{v}{q}.
\label{eq:goal_term}
\end{equation}

The secondary cost terms are
{\small
\begin{align}
\distTerm{t}
&:=
\sum_{q\in\agentSet}
\sum_{\orEdgeWrap{i}{j}\in\routingArcs{t}}
\travelDist{t}{\edgeWrap{i}{j}}
\arcSelectVar{t}{i,j}{q},
\\
\arcTerm{t}
&:=
\sum_{q\in\agentSet}
\sum_{\orEdgeWrap{i}{j}\in\routingArcs{t}}
\left(
1-\condEdgeExistProb{t}{\edgeWrap{i}{j}}
\right)
\arcSelectVar{t}{i,j}{q},
\\
\visitTerm{t}
&:=
\sum_{q\in\agentSet}
\sum_{v\in\candidateNodes{t}{q}}
\nodeVisitTimes{t}{v}
\multiAgentNodeSelectVar{t}{v}{q}.
\end{align}
}
Since grounded edges have unit existence probability, $\arcTerm{t}$ penalizes only the use of hypothesized connections. For $s\in\{\mathrm{dist},\mathrm{arc},\mathrm{visit}\}$, let $L^s$ and $U^s$ denote valid lower and upper bounds on the corresponding cost term. Each term $X$ is normalized as
\begin{equation}
\normTerm{t}{s}{X}
:=
\begin{cases}
0, & U^s=L^s,\\
\dfrac{X-L^s}{U^s-L^s}, & U^s>L^s.
\end{cases}
\label{eq:objective_normalization}
\end{equation}

Let $\pathLenPenalWeight$, $\arcUnreliaWeight$, and $\nodeRevisitWeight$ be nonnegative weights for normalized travel distance, edge uncertainty, and revisits, respectively. The resulting secondary route cost is
\begin{equation}
\begin{aligned}
\routeCostTerm{t}:={}&
\pathLenPenalWeight
\normTerm{t}{\mathrm{dist}}{\distTerm{t}}
+
\arcUnreliaWeight
\normTerm{t}{\mathrm{arc}}{\arcTerm{t}}
\\[-1mm]
&+
\nodeRevisitWeight
\normTerm{t}{\mathrm{visit}}{\visitTerm{t}}.
\end{aligned}
\label{eq:secondary_route_cost}
\end{equation}

We use a lexicographic objective because target discovery defines task success, while travel distance, edge uncertainty, and revisits determine how efficiently that success is achieved. The optimizer therefore first maximizes target reward and then minimizes routing cost without allowing efficiency gains to reduce the best achievable search reward. This also avoids introducing a cross-objective weighting parameter whose value could arbitrarily change the balance between search performance and route efficiency.

The lexicographic objective is
\begin{subequations}
\label{eq:lexicographic_objective}
\begin{align}
\min_{(x,y,u,a)\in\feasibleSet{t}}
\quad
&\routeCostTerm{t}
\label{eq:lexicographic_objective_secondary}
\\
\text{s.t.}\quad
&\goalTerm{t}=\optimalGoalTerm{t},
\nonumber
\end{align}
\end{subequations}
where $\optimalGoalTerm{t}:=\max_{(x,y,u,a)\in\feasibleSet{t}}\goalTerm{t}$.

The two levels are solved sequentially by first obtaining $\optimalGoalTerm{t}$ and then minimizing $\routeCostTerm{t}$ subject to $\goalTerm{t}=\optimalGoalTerm{t}$. Thus, the formulation is a hierarchical multi-objective model: the first level determines the reward-optimal solution set, while the second selects the most favorable route within that set.

\paragraph{Path Construction Constraints}
The planned route for each agent $q\in\agentSet$ is modeled as an open path rooted at its current viewpoint $\orCurrentVp{q}$. Each agent must depart exactly once from its current viewpoint:
\begin{equation}
\sum_{\orEdgeWrap{\orCurrentVp{q}}{j}\in\routingArcs{t}}
\arcSelectVar{t}{\orCurrentVp{q},j}{q}
=
1,
\quad
\forall q\in\agentSet .
\label{eq:multi_agent_depart}
\end{equation}

Moreover, only grounded edges can be executed immediately:
\begin{equation}
\sum_{j\in\viewpoints{1:t}:\edgeWrap{\orCurrentVp{q}}{j}\in\VPEdgesGrounded{1:t}}
\arcSelectVar{t}{\orCurrentVp{q},j}{q}
=
1,
\quad
\forall q\in\agentSet .
\label{eq:multi_agent_firsthop_grounded}
\end{equation}

The path is open and cannot return to the current viewpoint within the same planning round:
\begin{equation}
\sum_{\orEdgeWrap{i}{\orCurrentVp{q}}\in\routingArcs{t}}
\arcSelectVar{t}{i,\orCurrentVp{q}}{q}
=
0,
\quad
\forall q\in\agentSet .
\label{eq:no_return_current}
\end{equation}

For each candidate node, node selection is linked to incoming path flow:
\begin{equation}
\sum_{\orEdgeWrap{j}{n}\in\routingArcs{t}}
\arcSelectVar{t}{j,n}{q}
=
\multiAgentNodeSelectVar{t}{n}{q},
\quad
\forall n\in\candidateNodes{t}{q},~q\in\agentSet .
\label{eq:multi_agent_flowin}
\end{equation}

Each selected node can have at most one outgoing arc:
\begin{equation}
\sum_{\orEdgeWrap{n}{j}\in\routingArcs{t}}
\arcSelectVar{t}{n,j}{q}
\le
\multiAgentNodeSelectVar{t}{n}{q},
\quad
\forall n\in\candidateNodes{t}{q},~q\in\agentSet .
\label{eq:multi_agent_flowout}
\end{equation}

Let $\deg_{\VPEdgesGrounded{1:t}}(v)$ denote the number of physically verified navigable edges incident to node $v$. We define $\deadEndNodes{t}:=\left\{v\in\viewpoints{1:t}: \nodeVisitTimes{t}{v}>0,~\deg_{\VPEdgesGrounded{1:t}}(v)=1\right\}$.
These nodes correspond to previously visited dead ends. Revisiting such a node cannot reveal a new grounded continuation, because its only verified connection leads back to the already explored part of the graph. Therefore, to avoid unproductive backtracking, we exclude these nodes from planned paths:
\begin{equation}
\multiAgentNodeSelectVar{t}{v}{q}=0,
\quad
\forall q\in\agentSet,~
v\in\deadEndNodes{t}\cap\candidateNodes{t}{q}.
\label{eq:exclude_visited_dead_ends}
\end{equation}

The following constraint ensures that target-assignment variables can be activated only at selected reward endpoints:
\begin{equation}
\targetAssignVar{t}{k}{v}{q}
\le
\multiAgentNodeSelectVar{t}{v}{q},
\quad
\forall k\in\targetSet,~
v\in\rewardVPs{t}{q},~
q\in\agentSet .
\label{eq:target_assignment_visit}
\end{equation}

Moreover, if a reward coefficient is zero, the corresponding target-assignment variable is fixed to zero:
\begin{equation}
\targetAssignVar{t}{k}{v}{q}=0,
\quad
\forall k\in\targetSet,~
v\in\rewardVPs{t}{q},~
q\in\agentSet:
\targetProb{t}{v}{k}=0 .
\label{eq:target_assignment_positive}
\end{equation}

Each target can contribute reward at most once across the team, and a found target cannot contribute additional reward:
\begin{equation}
\sum_{q\in\agentSet}
\sum_{v\in\rewardVPs{t}{q}}
\targetAssignVar{t}{k}{v}{q}
\le
1-\targetFound{t}{k},
\quad
\forall k\in\targetSet .
\label{eq:target_assignment_unique}
\end{equation}

If $|\actTargets{t}|\ge|\agentSet|$ and the positive-reward agent--target graph $\{(q,k):\exists v\in\rewardVPs{t}{q},~\targetProb{t}{v}{k}>0\}$ admits a matching that covers all agents, we additionally impose
\begin{equation}
\sum_{k\in\actTargets{t}}
\sum_{v\in\rewardVPs{t}{q}}
\targetAssignVar{t}{k}{v}{q}
\ge
1,
\quad
\forall q\in\agentSet .
\label{eq:target_directed_agent_assignment}
\end{equation}

To prevent multiple agents from choosing the same immediate next viewpoint, we impose
\begin{equation}
\sum_{q\in\agentSet:\edgeWrap{\orCurrentVp{q}}{v}\in\VPEdgesGrounded{1:t}}
\arcSelectVar{t}{\orCurrentVp{q},v}{q}
\le
1,
\quad
\forall v\in\viewpoints{1:t}.
\label{eq:first_hop_multi_agent_unique}
\end{equation}

\paragraph{\textbf{Subtour Elimination}}
The above constraints enforce local path consistency but do not by themselves eliminate disconnected cycles. To suppress such subtours, we use Miller-Tucker-Zemlin constraints \cite{miller1960integer}. For all $n\in\orNodes{t}$ and $q\in\agentSet$, let $1\le\multiAgentSubtourVar{t}{n}{q}\le|\orNodes{t}|$.
For all $\orEdgeWrap{i}{j}\in\routingArcs{t}$ and $q\in\agentSet$, impose
\begin{equation}
\multiAgentSubtourVar{t}{i}{q}
-
\multiAgentSubtourVar{t}{j}{q}
+
|\orNodes{t}|
\arcSelectVar{t}{i,j}{q}
\le
|\orNodes{t}|-1.
\label{eq:mtz}
\end{equation}
These constraints ensure that the selected arcs for each agent form one ordered path rooted at the current viewpoint.

\paragraph{\textbf{Normalization Bounds}}
Let $\firstHopEdges{t}{q}:=\{j\in\viewpoints{1:t}:
\edgeWrap{\orCurrentVp{q}}{j}\in\VPEdgesGrounded{1:t}\}$, $d_{\max}^{t}:=\max_{\orEdgeWrap{i}{j}\in\routingArcs{t}}
\travelDist{t}{\edgeWrap{i}{j}}$,
and
$r_{\max}^{t}:=\max_{\orEdgeWrap{i}{j}\in\routingArcs{t}}
(1-\condEdgeExistProb{t}{\edgeWrap{i}{j}})$.
The bounds in \eqref{eq:objective_normalization} are
{\small
\begin{gather}
    \pathLenPenalWeightLB
    =
    \sum_{q\in\agentSet}
    \min_{j\in\firstHopEdges{t}{q}}
    \travelDist{t}{\edgeWrap{\orCurrentVp{q}}{j}},
    \\
    \pathLenPenalWeightUB
    =
    \sum_{q\in\agentSet}
    \left[
    \max_{j\in\firstHopEdges{t}{q}}
    \travelDist{t}{\edgeWrap{\orCurrentVp{q}}{j}}
    +
    (|\orNodes{t}|-2)d_{\max}^{t}
    \right],
    \\
    \arcUnreliaWeightLB
    =0,\qquad
    \arcUnreliaWeightUB
    =
    |\agentSet|(|\orNodes{t}|-1)r_{\max}^{t},
    \\
    \nodeRevisitWeightLB
    =
    \sum_{q\in\agentSet}
    \min_{j\in\firstHopEdges{t}{q}}
    \nodeVisitTimes{t}{j}, \qquad
    \nodeRevisitWeightUB
    =
    \sum_{q\in\agentSet}
    \sum_{v\in\candidateNodes{t}{q}}
    \nodeVisitTimes{t}{v}.
\end{gather}
}

These bounds follow from the grounded first-hop requirement and the open simple-path structure enforced by the MTZ constraints; the upper bounds use the corresponding maximum per-arc costs over at most $|\orNodes{t}|-1$ selected arcs.

\section{Experiments}
\label{sec:numerical_experiments}

We evaluate the proposed framework and comparison methods in the Matterport3D Simulator \cite{anderson2018vision}. The evaluation consists of 100 test cases drawn from five indoor scans in the Matterport3D dataset \cite{Matterport3D}. For each scan, target objects are manually selected and annotated with a textual description and the indices of viewpoints whose panoramas contain the target. These visibility annotations are used exclusively by the offline Oracle and for performance evaluation. The test cases span team sizes of $\{1,3,5\}$ agents and target-set sizes of $\{1,4,8\}$ objects, with randomized initial agent locations and target selections. Each online method is run for at most 30 decision steps and uses the same GPT-5.4 model \cite{openai2026gpt54}, configured with medium reasoning effort, to detect targets from panoramic observations.

\subsection{Baseline Methods}
\label{subsec:comparison_methods}

\paragraph{\textbf{Offline Perfect-Knowledge Oracle}}
The Oracle is given the complete navigation graph, exact edge distances, agent start nodes, and detectable-viewpoint set $\oracleDetectable{i}{k}$ for every target $k$ and test case $i$. It first computes shortest paths among the agent start nodes and all target-detectable viewpoints. An exact per-agent dynamic program then determines the minimum-distance walk associated with each attainable target-coverage set. A second dynamic program combines these per-agent walks to cover all targets while minimizing the total distance traveled by the team. Targets visible from intermediate nodes along a route are also counted as covered. The Oracle is unavailable to the online methods and provides the performance upper bound.

\paragraph{\textbf{VLFM-G}}
VLFM-G is our multi-agent extension of Vision-Language Frontier Maps (VLFM) \cite{yokoyama2024vlfm}. It ranks observed, unvisited frontier viewpoints by target-conditioned visual-semantic relevance and coordinates agents over the online grounded partial graph. For each frontier, eight directional panorama views are scored against target descriptions using the BLIP-2 image--text contrastive head \cite{pmlrv202li23q}, then fused with angular-confidence weighting to maintain its semantic value \cite{yokoyama2024vlfm}.
Let $\bar S_v^t$ denote the normalized maximum semantic value of frontier $v$ over all active targets, and let $\bar d_{qv}^t$ denote the normalized grounded shortest-path distance from agent $q$ to frontier $v$. For a nonnegative distance weight $\beta$, VLFM-G assigns agents to distinct frontiers by maximizing $U_{qv}^t
=\bar S_v^t
-
\beta \bar d_{qv}^t$ using the Hungarian algorithm \cite{kuhn1955hungarian}, with dummy assignments representing wait actions. Conflicts in the agents' first hops are resolved by retaining the highest-utility assignment and reassigning the remaining agents. Only the first grounded hop is executed before the next replanning step. The distance weight $\beta$ is calibrated and selected using 20 held-out episodes that are disjoint from the test set.

\paragraph{\textbf{Decentralized No-Sharing Variant (Dec-Graph)}}
Dec-Graph is a controlled variant of the proposed framework in which agents maintain separate graphs and select actions independently, while retaining the same target detector, hypothesis-update mechanism, and optimization model. Each agent $q\in\agentSet$ independently uses MLLM to construct and update a private graph $\graph{1:t}_{q}$ from only its own observation history and then solves the single-agent specialization of \eqref{eq:lexicographic_objective} with $\agentSet=\{q\}$.
For a fair comparison across methods, the target-found status is the only information shared among agents: whenever an agent detects and claims a target as found, this status is communicated to all other agents, which subsequently also mark that target as found and remove it from their active target sets. Apart from this synchronized target-found status, no graph, observation, hypothesis, route, or action information is exchanged among agents. Consequently, target assignments and first-hop choices are not jointly coordinated: different agents may pursue the same active target or simultaneously occupy the same viewpoint while retaining separate observation histories and graph states. Dec-Graph coincides with the proposed method when $|\agentSet|=1$ and is therefore reported only for the multi-agent cases.

\subsection{Evaluation Metrics}
\label{subsec:evaluation_metrics}

For each episode $i\in\episodeSet$, let $\evalAgents{i}\subseteq\agentSet$ be the robot team, $\evalTargets{i}\subseteq\targetSet$ the target set, $\evalViewpoints{i}$ the complete evaluation navigation viewpoint set, and $\episodeHorizon{i}$ the terminal time. Let $\evalCurrentVp{i}{t}{q}\in\evalViewpoints{i}$ denote the viewpoint occupied by robot $q$ at time $t$ in episode $i$. For target $k$, let $\detOutput{i}{t}{q}{k}=1$ indicate that robot $q$ claims to detect target $k$ at time $t$ in episode $i$.
A target is counted as verified only when a claimed detection is made from an oracle-detectable viewpoint, i.e., $\evalCurrentVp{i}{t}{q}\in\oracleDetectable{i}{k}$. Specifically, we define $\verifiedFound{i}{k}=1$ if there exist $t\leq\episodeHorizon{i}$ and $q\in\evalAgents{i}$ such that $\detOutput{i}{t}{q}{k}=1$ and $\evalCurrentVp{i}{t}{q}\in\oracleDetectable{i}{k}$, which means that at least one robot reports detecting target $k$ while occupying a viewpoint from the associated oracle-detectable viewpoints, and 0 otherwise.

\paragraph{\textbf{Progress}}
For multi-target search, we additionally report the fraction of verified targets found in each episode as
$
\episodeProgress{i}
:=
\frac{1}{|\evalTargets{i}|}
\sum_{k\in\evalTargets{i}}
\verifiedFound{i}{k}.
$
Then the dataset-level Progress score is
$
\progressScore
:=
\frac{1}{|\episodeSet|}
\sum_{i\in\episodeSet}
\episodeProgress{i},
$
which assigns partial credit when a method finds only a subset of the requested targets.

\paragraph{\textbf{Team-PPL}}
To jointly evaluate search efficiency and partial task completion, we use team-level Progress weighted by Path Length. Let $\agentTravelCost{i}{q}$ denote the distance traveled by robot $q$ in episode $i$. The total team distance and makespan distance are defined as
$
\totalTeamDist{i}
:=
\sum_{q\in\evalAgents{i}}
\agentTravelCost{i}{q},
$
and
$
\makespanTeamDist{i}
:=
\max_{q\in\evalAgents{i}}
\agentTravelCost{i}{q},
$
respectively. The Oracle minimizes total team distance. Accordingly, let $\totalTeamDistOracle{i}$ and $\makespanTeamDistOracle{i}$ denote the total team distance and makespan distance within the Oracle solution for episode $i$, respectively. We define
\begin{equation}
\teamPPLTotal
:=
\frac{1}{|\episodeSet|}
\sum_{i\in\episodeSet}
\episodeProgress{i}
\frac{\totalTeamDistOracle{i}}
{\max\left(
\totalTeamDist{i},
\totalTeamDistOracle{i}
\right)},
\end{equation}
which evaluates the trade-off between verified target coverage and aggregate team travel. A higher $\teamPPLTotal$ indicates that the team finds a larger fraction of the targets while keeping its total travel distance closer to the perfect-information Oracle.

We further define
\begin{equation}
\teamPPLMakespan
:=
\frac{1}{|\episodeSet|}
\sum_{i\in\episodeSet}
\episodeProgress{i}
\frac{\makespanTeamDistOracle{i}}
{\max\left(
\makespanTeamDist{i},
\makespanTeamDistOracle{i}
\right)},
\end{equation}
which evaluates verified target coverage relative to the longest route traveled by any individual robot. A higher $\teamPPLMakespan$ indicates that the team achieves greater search progress while limiting its longest individual route and therefore reflects more efficient parallel execution.

\paragraph{\textbf{Route Detection F1 Score}}
We report detection F1 as a diagnostic of detector performance along each method's executed trajectories. For each detection query, $\detOutput{i}{t}{q}{k}$ is evaluated against whether $\evalCurrentVp{i}{t}{q}\in\oracleDetectable{i}{k}$. Since different methods visit different viewpoints and find targets at different times, they induce different detection-query sets; hence, their F1 scores are not directly comparable.

\paragraph{\textbf{Probability-Weighted Geodesic Score (PWGS)}}
PWGS measures how well the predicted target-location distribution places probability mass near viewpoints from which the target is actually detectable. For target $k$ in episode $i$, let $\oracleGeodesicDist{i}{v}{u}$ be the ground-truth geodesic distance between viewpoints $v$ and $u$, $\targetOracleDist{i}{k}{v}:=\min_{u\in\oracleDetectable{i}{k}}\oracleGeodesicDist{i}{v}{u}$ the distance from $v$ to the nearest oracle-detectable viewpoint for target $k$, and $\normTargetOracleDist{i}{k}{v}:=\targetOracleDist{i}{k}{v}/\max_{u\in\evalViewpoints{i}}\targetOracleDist{i}{k}{u}$ its normalized value. At time $t$, the score for active target $k$ is $1-\sum_{v\in\eligibleVPs{t}}\targetProb{t}{v}{k}\normTargetOracleDist{i}{k}{v}$. A score of $1$ means that all probability mass is assigned to oracle-detectable viewpoints, while lower scores indicate that more probability is assigned farther from where the target can actually be detected. Within each episode, PWGS is averaged over all active target--time pairs, and the resulting episode-level scores are then averaged across episode set $\episodeSet$. In our evaluation, the generated hypothesized edge candidates did not satisfy the required admissibility checks and were therefore removed before graph materialization, while the remaining graph components passed validation and were retained. Consequently, no edge hypothesis were available for evaluating edge-existence or edge-distance accuracy, and we do not report a separate edge-hypothesis metric. This reflects a limitation of current MLLMs in reliably generating valid edge hypotheses within the joint graph-generation task.

\paragraph{\textbf{Implementation details}}
All proposed MLLM variants use identical graph-update and planning configurations. Following calibration, the target-directed MILP employs the two-priority lexicographic objective in \eqref{eq:lexicographic_objective}, with the secondary-level weights set to $(\pathLenPenalWeight,\arcUnreliaWeight,\nodeRevisitWeight)=(0.17,0.05,0.30)$. For every newly hypothesized edge $\edgeWrap{i}{j}\in\VPEdgesUngrounded{t}$, the initial distance variance is set to $\MLLMdistVarVV{t}{\edgeWrap{i}{j}}=4.0$. The remaining hypothesis-update parameters are $\VVedgeCueFactor=1.0$, $\edgeExistLikelihoodFactor=0.7$, and $\smallNumber=10^{-6}$. The parameter $\beta$ is selected from $\{0,0.1,0.25,0.5,1\}$ using 20 held-out calibration episodes, with $\beta=0$ yielding the highest $\teamPPLTotal$ and therefore being used in all reported experiments.
Dec-Graph uses the instruction-tuned Gemma 4 31B model in non-thinking mode \cite{gemmateam2026gemma4} for hypothesis graph generation. It uses the same hypothesis-update and optimization parameters, but performs an independent graph update and MILP solve for each agent.
Finally, all MILPs are solved using Gurobi~12.0.1 with a time limit of 30 seconds.

\subsection{Experiment Results}
\label{subsec:exp_results}
The four proposed variants differ only in the graph-generation MLLM or inference mode. GPT-Medium uses GPT-5.4 with medium reasoning effort \cite{openai2026gpt54}; Gemma-Instant uses instruction-tuned Gemma 4 31B without thinking \cite{gemmateam2026gemma4}; and Qwen-Thinking and Qwen-Instant use Qwen3.6-35B-A3B with thinking enabled and disabled, respectively \cite{qwen36_35b_a3b}. Target detection, graph updating, optimization, execution, and episode budgets are identical across variants.

All online methods use the same 100 episodes, initial viewpoints, target descriptions, detector, local navigation interface, and 30-step budget. Detection precedes action selection, and agents execute at most one synchronous local hop per step. Tables~\ref{tab:single_agent_results} and \ref{tab:multi_agent_results} report the 34 single-agent episodes ($|\evalAgents{i}|=1$) and 66 multi-agent episodes ($|\evalAgents{i}|\in\{3,5\}$) (i.e., 3 and 5 robots), respectively; both subsets include $|\evalTargets{i}|\in\{1,4,8\}$ (i.e., 1,4,and 8 targets to be found). Bold and underlined entries mark the best and second-best online results, with ties marked jointly and the offline Oracle excluded from ranking.
\begin{table}[tbp]
\small
\centering
\caption{Performance on the single-agent test cases (* indicates the benchmark method).}
\label{tab:single_agent_results}
\setlength{\tabcolsep}{4pt}
\renewcommand{\arraystretch}{1.12}
\resizebox{0.42\textwidth}{!}{%
\begin{tabular}{@{}lccccc@{}}
\toprule
\multicolumn{1}{l}{\multirow{2}{*}{Method}}
& \multicolumn{1}{c}{\multirow{2}{*}{Progress}}
& \multicolumn{2}{c}{Team-PPL}
& \multicolumn{1}{c}{\multirow{2}{*}{F1}}
& \multicolumn{1}{c}{\multirow{2}{*}{PWGS}} \\
\cmidrule(lr){3-4}
&
&
\multicolumn{1}{c}{Total}
& \multicolumn{1}{c}{Makespan}
&
& \\
\midrule
Oracle
& 1.000
& 1.000
& 1.000
& --
& - \\

GPT-Medium
& \underline{0.271}
& \textbf{0.172}
& \textbf{0.172}
& \underline{0.332}
& \textbf{0.601} \\

Qwen-Thinking
& \textbf{0.275}
& 0.146
& 0.146
& 0.330
& \underline{0.590} \\

Qwen-Instant
& \underline{0.271}
& 0.142
& 0.142
& \textbf{0.388}
& 0.568 \\

Gemma-Instant
& 0.217
& \underline{0.164}
& \underline{0.164}
& 0.310
& 0.580 \\

VLFM-G*
& 0.258
& 0.133
& 0.133
& 0.279
& -- \\
\bottomrule
\end{tabular}%
}
\end{table}

\paragraph{\textbf{Single-agent search}}
The single-agent results demonstrate a consistent advantage of the proposed graph-based search formulation over VLFM-G, with Gemma-Instant representing the only exception in terms of Progress. GPT-Medium provides the clearest illustration: relative to VLFM-G, Progress increases $5\%$, while Team-PPL improves $29\%$. Gemma-Instant achieves lower Progress despite maintaining competitive efficiency and hypothesis-quality scores. This result is substantially influenced by six episodes that terminated prematurely following repeated failures to generate schema-valid graph updates.
A second trend is that additional reasoning during graph generation improves the information provided to the planner. GPT-Medium ranks first or second across all reported single-agent metrics. The controlled Qwen comparison supports this trend: enabling thinking yields relative improvements of $1.5\%$ in Progress, $2.8\%$ in Team-PPL, and $3.9\%$ in PWGS. These consistent gains suggest that additional reasoning produces more useful spatial hypotheses for downstream search.

\begin{table}[tbp]
\small
\centering
\caption{Performance on the multi-agent test cases (* indicates benchmark methods).}
\label{tab:multi_agent_results}
\setlength{\tabcolsep}{4pt}
\renewcommand{\arraystretch}{1.12}
\resizebox{0.42\textwidth}{!}{%
\begin{tabular}{@{}lccccc@{}}
\toprule
\multicolumn{1}{l}{\multirow{2}{*}{Method}}
& \multicolumn{1}{c}{\multirow{2}{*}{Progress}}
& \multicolumn{2}{c}{Team-PPL}
& \multicolumn{1}{c}{\multirow{2}{*}{F1}}
& \multicolumn{1}{c}{\multirow{2}{*}{PWGS}} \\
\cmidrule(lr){3-4}
&
& \multicolumn{1}{c}{Total}
& \multicolumn{1}{c}{Makespan}
&
& \\
\midrule
Oracle
& 1.000
& 1.000
& 1.000
& --
& - \\

GPT-Medium
& \textbf{0.500}
& \underline{0.129}
& \underline{0.267}
& \textbf{0.449}
& \textbf{0.697} \\

Qwen-Thinking
& 0.465
& \textbf{0.133}
& \textbf{0.290}
& \underline{0.431}
& 0.629 \\

Qwen-Instant
& \underline{0.474}
& 0.105
& 0.244
& 0.379
& 0.596 \\

Gemma-Instant
& 0.467
& 0.102
& 0.233
& 0.401
& \underline{0.636} \\

Dec-Graph*
& 0.403
& 0.089
& 0.169
& 0.148
& -- \\

VLFM-G*
& 0.311
& 0.076
& 0.157
& 0.077
& -- \\
\bottomrule
\end{tabular}%
}
\vspace{-0.5cm}
\end{table}

\begin{figure}[t]
    \centering
    \begin{minipage}{0.45\textwidth}
        \centering
        \includegraphics[width=0.9\linewidth]{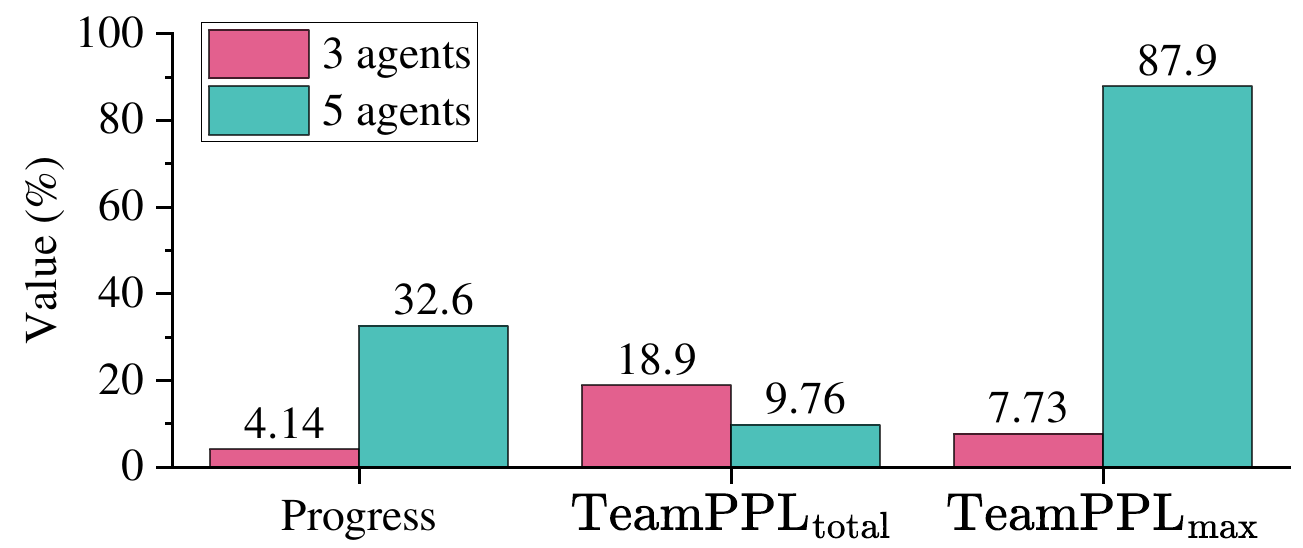}
        \caption{Relative improvement of Gemma-Instant from Dec-Graph.}
        \label{fig:bar_plot}
    \end{minipage}
\end{figure}

\paragraph{\textbf{Multi-agent search}}
The advantage of the proposed framework becomes more pronounced in the multi-agent setting. All MLLM variants outperform the benchmark methods across every reported metric, with a substantially clearer margin than in the single-agent setting. This indicates that the proposed hypothesis-graph formulation benefits particularly from cooperative search, where shared information can be exploited across multiple agents.

The choice of MLLM also has a strong effect under multi-agent search. In particular, the PWGS range across MLLM variants increases from $0.033$ in the single-agent setting to $0.101$ in the multi-agent setting. This wider spread suggests that MLLM capability becomes consequential as the planner must construct spatial hypotheses that support coordinated decisions across multiple agents.

To illustrate how the benefits of cooperation vary with team size, \autoref{fig:bar_plot} compares the three- and five-agent settings. Relative to the three-agent team, the five-agent team achieves larger improvements over its decentralized counterpart (Dec-Graph) in progress (from 4.14\% to 32.60\%) and maximum Team-PPL (from 7.73\% to 87.9\%)), indicating that cooperative navigation becomes more advantageous as the team expands.

\begin{figure*}[ht]
    \centering
    \begin{minipage}{1\textwidth}
        \centering
        \includegraphics[width=\linewidth]{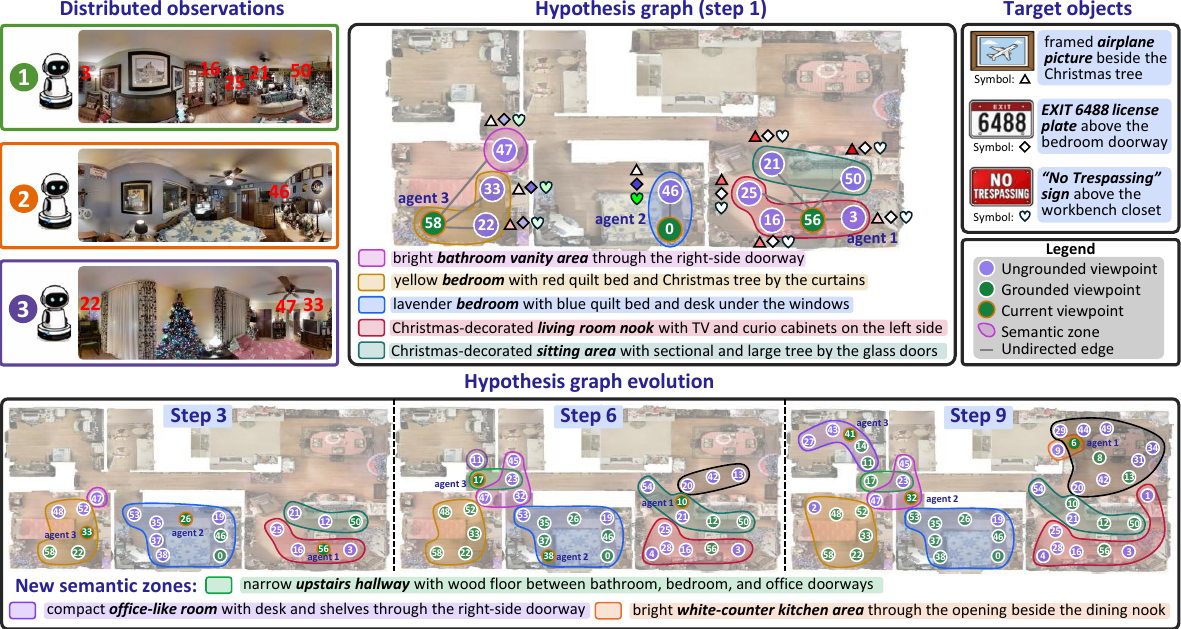}
        \caption{Illustrative GPT-Medium episode. Darker target symbols indicate higher estimated detectability; the floor plan is shown only for visualization and is unavailable to the agents.}
        \label{fig:test_case}
    \end{minipage}
    \vspace{-0.3cm}
\end{figure*}

\begin{figure*}[t]
    \centering
    \begin{minipage}{\textwidth}
        \centering
        \includegraphics[width=\linewidth]{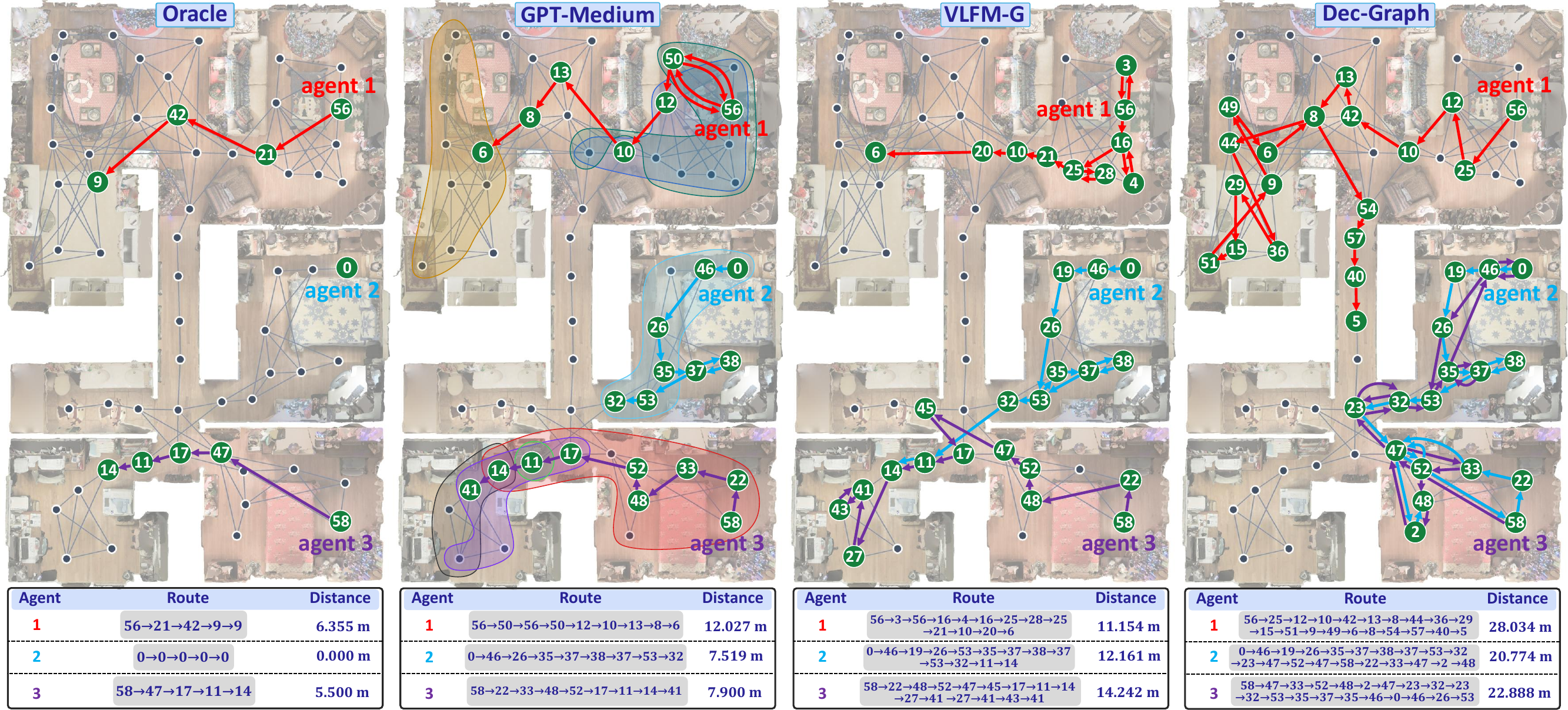}
        \caption{Final routes for the Oracle, GPT-Medium, VLFM-G, and Dec-Graph. The tables report each agent's viewpoint sequence and travel distance.}
        \label{fig:final_routes}
    \end{minipage}
    \vspace{-0.2cm}
\end{figure*}

A detailed illustration of the online behavior is provided in \autoref{fig:test_case}, which presents one GPT-Medium episode with eight targets, three of which are shown for clarity. In the figure, target-symbol shape identifies the target, and darker symbols indicate higher estimated probability that the corresponding target locates at the current viewpoint. We can observe that the inferred semantic zones align well with the agents' observations, and the estimated target-location probabilities remain consistent with the observed environmental context. For example, the strong airplane-picture hypotheses at viewpoints 21 and 50 are consistent with the description “beside the Christmas tree”. The other targets are likewise distributed over semantically plausible viewpoints.

\autoref{fig:final_routes} further compares the resulting routes using the same episode for illustration. In this case, it can be observed that GPT-Medium distributes the exploration workload relatively evenly among the three agents. Compared with VLFM-G and Dec-Graph, it reduces total travel distance by $26.9\%$ and $61.7\%$, and reduces makespan by $15.6\%$ and $57.1\%$. The route patterns in the figure further show that these gains arise from shared-graph coordination and cooperative navigation, which directs agents toward complementary branches and avoids the repeated exploration observed with Dec-Graph.

\section{Conclusion}
This paper presented a training-free and uncertainty-aware framework for cooperative multi-agent multi-target embodied search in mapless, partially observed environments. MLLMs are used to generate structured hypotheses over semantic regions, target locations, and navigable connectivity, which are progressively refined in a shared multi-resolution graph. A rolling-horizon MILP then converts these uncertain hypotheses into coordinated search decisions, balancing target discovery, travel efficiency, uncertainty, and repeated exploration.

The framework was evaluated on 100 Matterport3D \cite{Matterport3D} test cases using multiple MLLM backbones \cite{openai2026gpt54,gemmateam2026gemma4,qwen36_35b_a3b} and compared with VLFM-G (a multi-agent VLFM \cite{yokoyama2024vlfm} adaptation), Dec-Graph (a decentralized baseline), and a perfect-knowledge Oracle. The results show consistent improvements over the benchmark methods in search progress and travel efficiency, with the advantage becoming more pronounced in multi-agent settings. The target-location results further show that MLLM reasoning quality affects how strongly the predicted probability mass is concentrated near the true targets in geodesic distance, particularly in cooperative search. Moreover, comparison with Dec-Graph demonstrates that shared graph information and joint planning substantially improve coordination, especially by balancing workloads and enabling more effective parallel exploration.

Performance is mainly limited by the accuracy and calibration of target-location and connectivity hypotheses and by the models’ visual grounding. Future work will incorporate stronger perception and verification, online confidence calibration, and active hypothesis validation to approach Oracle-level performance. We will also evaluate larger environments, agent teams, and dynamic scenes, followed by physical-robot experiments under realistic sensing, localization, communication, and execution uncertainty. These advances will support scalable and reliable real-world cooperative embodied systems.

\bibliographystyle{IEEEtran}
\bibliography{refs}

\end{document}